\documentclass[letterpaper, 10 pt, journal, twoside]{IEEEtran}

\IEEEoverridecommandlockouts
\usepackage{graphics}
\usepackage{epsfig}
\usepackage{mathptmx}
\usepackage{times}
\usepackage{amsmath}
\usepackage{amssymb}
\usepackage{ulem}
\usepackage{hyperref}

\usepackage{xcolor}
\usepackage[dvipsnames]{xcolor}
\newcommand{\redcross}{{\color{Red}$\times$}}
\newcommand{\greencheck}{{\color{ForestGreen}\checkmark}}
\newcommand{\citet}{\cite}

\usepackage{tikz}
\usepackage{textcomp}
\usepackage{hyperref}
\usepackage{lipsum}
\newcommand\copyrighttext{%
  \footnotesize \textcopyright 2025 IEEE. Personal use of this material is permitted.
  Permission from IEEE must be obtained for all other uses, in any current or future 
  media, including reprinting/republishing this material for advertising or promotional 
  purposes, creating new collective works, for resale or redistribution to servers or 
  lists, or reuse of any copyrighted component of this work in other works. Link to published paper: \href{https://ieeexplore.ieee.org/document/11222773}{https://ieeexplore.ieee.org/document/11222773}. DOI: 10.1109/LRA.2025.3627093}
\newcommand\copyrightnotice{%
\begin{tikzpicture}[remember picture,overlay]
\node[anchor=north,yshift=-10pt] at (current page.north) {\fbox{\parbox{\dimexpr\textwidth-\fboxsep-\fboxrule\relax}{\copyrighttext}}};
\end{tikzpicture}%
}

\begin{document}

\markboth{IEEE Robotics and Automation Letters. Preprint Version. Accepted Oct, 2025} {Kwan \MakeLowercase{\textit{et al.}}: Onboard Mission Replanning for Adaptive Cooperative Multi-Robot Systems} 

\author{Elim Kwan$^{1}$, Rehman Qureshi$^{2}$, Liam Fletcher$^{1}$, Colin Laganier$^{1}$, Victoria Nockles$^{1}$, Richard Walters$^{1}$
\thanks{Manuscript received: June, 2, 2025; Revised September, 1, 2025; Accepted October, 5, 2025. This article was recommended for publication by Associate Editor J. Rojas and Editor J. Kober upon evaluation of the reviewers' comments. This work was supported by the Alan Turing Institute's Defence and National Security programme through a partnership with Dstl. This work was partially supported by an international internship on behalf of the Science, Mathematics, and Research for Transformation (SMART) Scholarship-for-Service Program within the OUSD/R\&E (The Under Secretary of Defense-Research and Engineering), National Defense Education Program (NDEP) / BA-1, Basic Research. We used the Baskerville Tier 2 HPC, funded by EPSRC and UKRI (EP/T022221/1, EP/W032244/1), operated by Advanced Research Computing at the University of Birmingham.}
\thanks{$^{1}$The Alan Turing Institute, British Library, 96 Euston Rd., London NW1 2DB, United Kingdom
         {\tt\small ekwan@turing.ac.uk}}%
\thanks{$^{2}$Department of Aerospace Engineering, Auburn University, Auburn, AL 36849, United States
         {\tt\small rsq0001@auburn.edu}}%
\thanks{Code available at: \href{https://github.com/alan-turing-institute/graph-attention-replanner}{https://github.com/alan-turing-institute/graph-attention-replanner}}
\thanks{Digital Object Identifier (DOI): 10.1109/LRA.2025.3627093}
}

\title{Onboard Mission Replanning for \protect\\ Adaptive Cooperative Multi-Robot Systems
}

\maketitle
\copyrightnotice

\thispagestyle{empty}

\begin{abstract}
    Cooperative autonomous robotic systems have significant potential for executing complex multi-task missions across space, air, ground, and maritime domains. But they commonly operate in remote, dynamic and hazardous environments, requiring rapid in-mission adaptation without relying on fragile or slow communication links to centralized compute. Fast, on-board replanning algorithms are therefore essential to enhance resilience for these systems, but do not yet exist. 
    Reinforcement Learning (RL) shows strong promise for efficiently solving mission planning tasks formulated as Travelling Salesperson Problems (TSPs), but existing methods: 1) are unsuitable for replanning, where agents do not start at a single location; 2) do not allow cooperation between agents; 3) are unable to model tasks with variable durations; or 4) lack practical considerations for on-board deployment.
    Here we address this gap by defining the Cooperative Mission Replanning Problem as a novel adaptation of multiple TSP, and develop a new encoder/decoder-based RL model to solve it effectively and efficiently.
    Using a simple example of cooperative drones, we show our replanner consistently (90\% of the time) maintains performance within 10\% of the state-of-the-art LKH3 heuristic solver, whilst running 85-370 times faster on a Raspberry Pi. This work paves the way for increased resilience in autonomous multi-agent systems.
\end{abstract}

\begin{IEEEkeywords}
Path Planning for Multiple Mobile Robots or Agents, Task Planning, Multi-Robot Systems, AI and Machine Learning in Manufacturing and Logistics Systems, Aerial Systems: Perception and Autonomy
\end{IEEEkeywords}

\section{Introduction}
\label{sec:introduction}
\begin{figure}[!t]
\centering
\includegraphics[width=0.5\textwidth]{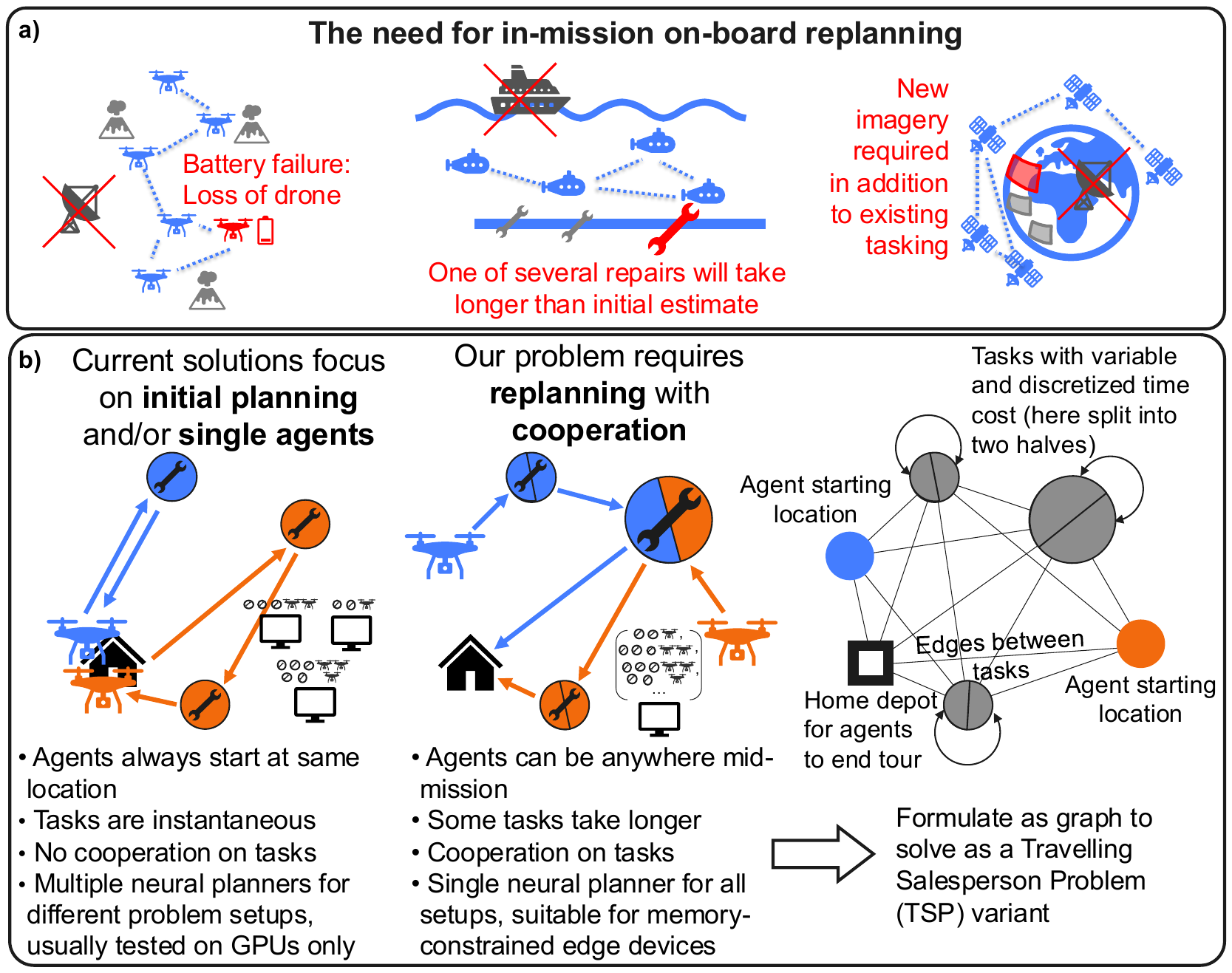}
\caption{(a) A cartoon illustrating the need for in-mission, on-board replanning through three examples that all feature major unforeseen changes and limited ground station communication in dynamic, challenging environments. (b) A cartoon illustrating differences between a more traditional initial planning, non-cooperative problem (left) and our cooperative replanning problem (middle). The right-hand diagram shows how this replanning problem can be formulated in a graph format as a variant of the multiple Travelling Salesperson Problem (mTSP).}
\label{fig:intro}
\vspace{-0.8cm}
\end{figure}

\IEEEPARstart{T}{eams} of autonomous robots, including uncrewed aerial, underwater, surface, or ground vehicles (UAVs, UUVs, USVs, UGVs) and multi-satellite constellations, have huge potential for efficiently and safely completing a diverse range of complex missions. In comparison to single-agent operations, such multi-agent systems have increased robustness to failure \cite{safety_in_number}, greater areal coverage \cite{coverage}, and faster mission completion \cite{thakur_swarm_2021}. 
Furthermore, they can cooperatively tackle complex missions comprised of multiple tasks, where these tasks can be as diverse as capturing images \cite{satellite1}, manufacturing/ repairing items or transporting materials \cite{node_cost}, or making environmental measurements \cite{palma}. However, these systems also commonly operate in environments that are both remote or isolated and also highly dynamic and hazardous, e.g. in nuclear power plants \cite{powerengineering}, over active volcanoes \cite{palma} or in congested low earth orbits in space \cite{satellite1}. Operating in such challenging environments requires these systems to be able to adapt and replan in response to major unexpected changes that impact the mission plan \cite{rizk_cooperative_2020} such as the identification of new tasks or the loss or failure of a robotic agent, without relying on fragile or slow communication links to centralized ground stations \cite{live-fly} (Fig. \ref{fig:intro}a). Therefore, lightweight and fast replanning algorithms that can be deployed on-board cooperative robotic systems are critical for enabling autonomous adaptability during a mission; but these do not yet exist.

\begin{table*}[t]
    \caption{Comparing the literature on using Attention Mechanism and Reinforcement Learning for Travelling Salesperson Problems}
    \label{tab:gat_lit_table}
    \vspace{5pt}
    \resizebox{1\textwidth}{!}{%
    \begin{tabular}{cllll|ccc|cc}
        \hline
        \multicolumn{5}{c}{} & \multicolumn{3}{|c|}{Key Adaptation} & \multicolumn{2}{c}{Practical Consideration} \\\
 Paper & Problem & Encoder & Decoder & Algo & \begin{tabular}[c]{@{}c@{}}Vary\\Start\\Locs\end{tabular} & \begin{tabular}[c]{@{}c@{}}Vary\\Time\\Cost\end{tabular} & \begin{tabular}[c]{@{}c@{}}Collab.\\ Tasking\end{tabular} & Generalization & \begin{tabular}[c]{@{}c@{}}On Edge\\ Devices\end{tabular} \\
        \hline
        \citet{bello2017neuralcombinatorialoptimizationreinforcement} & TSP & RNN & Ptr. Net. & REINFORCE \hspace{100pt} & \redcross & \redcross & \redcross & \redcross & \redcross \\
        \citet{deudon} & TSP & Attention & Ptr. Net. & REINFORCE (Rollout) & \redcross & \redcross & \redcross & \redcross & \redcross \\
        \citet{hgpn} & TSP & RNN + Graph Attention & Ptr. Net. & REINFORCE (Rollout) & \redcross & \redcross & \redcross & \redcross & \redcross \\
        \citet{kool2019attentionlearnsolverouting} & CVRP & Graph Attention & Attention Model & REINFORCE (Rollout) & \redcross & \redcross & \redcross & \redcross & \redcross \\
        \citet{mvmoe} & multi-tasks VRP & Graph Attention & Attention Model & POMO + MoE Layer & \redcross & \redcross & \redcross & \greencheck (Goal) & \redcross \\
        \citet{shi_multi-agent_2023} & mTSP with Multiple Depots & Graph Attention & Ptr. Net. & REINFORCE & \greencheck & \redcross & \redcross & \redcross & \redcross \\
        \citet{ucl_multi_robot} & mTSP with Multiple Depots & MoE Net. & Ptr. Net. & Pretrained & \greencheck & \redcross & \redcross & \greencheck (Agent) & \greencheck \\
        \citet{nazari2018reinforcementlearningsolvingvehicle} & Split Delivery VRP & Convolutional & Ptr. Net. & REINFORCE & \redcross & \redcross & \greencheck & \redcross & \greencheck \\
        \citet{dpn} & VRP (Min-Max) & Graph Attention (Partition/Navigation) & Attention Model & REINFORCE (Rollout) & \greencheck & \redcross & \redcross & \greencheck(Depot) & \redcross \\
       \citet{parco} & Heterogeneous CVRP (Min-Max) & Graph Attention & Attention Model (Parallel) & REINFORCE & \greencheck & \redcross & \redcross & \greencheck (Agent) & \redcross \\
 This Study & CMRP & Graph Attention & Attention Model & REINFORCE (Rollout) & \textbf{\greencheck} & \textbf{\greencheck} & \textbf{\greencheck} &\begin{tabular}[c]{@{}c@{}} \greencheck (Task, Agent, Collab.) \end{tabular} & \textbf{\greencheck} \\
        \hline \\
        \multicolumn{10}{l}{\small VRP refers to the Vehicle Routing Problem; CVRP refers to the Capacitated VRP; Ptr. Net. refers to the Pointer Network; MoE refers to the Mixture of Experts technique;} \\
        \multicolumn{10}{l}{\small POMO is a modified REINFORCE algorithm which uses policy optimization with multiple optima.} \\
    \end{tabular}
}
\end{table*}

Because mission planning with multiple agents requires finding the optimal tour path for each agent while minimizing the length of the longest tour, previous studies have often formulated and solved it as a variant of the multiple Traveling Salesman Problem (mTSP) with a min-max objective \cite{mtsp}. However, our particular planning problem has three major differences to traditional mTSP (Fig. \ref{fig:intro}b), and in addition requires practical consideration for edge deployment onboard robotic systems. First, variable start locations for agents must be considered, as each agent may be at any location within the mission domain when replanning is required. Second, nodes with variable time costs are necessary to reflect the common real-world situation where mission time depends not just on how long it takes for an agent to travel to a task, but also how long it takes to complete that task. This models a wide range of applications where tasks have differing execution times~\cite{aggarwal_extended_2022, uav_remote_sensing, uav_data_collection}. Third, cooperative tasking is required, such that multiple agents can work together on a single task to complete it faster, even if this joint contribution is asynchronous rather than simultaneous. No previous studies have addressed this important combination of multiple adaptations (see ``Key Adaptation'' columns of Table \ref{tab:gat_lit_table}), which we name the Cooperative Mission Replanning Problem (CMRP). Finally, to enable onboard replanning, any model for solving CMRP must be suitably small and fast \cite{elim_bnn}. These limitations also necessitate a single generalized model that can solve a range of problem sizes, as the number of available agents and remaining tasks are likely to vary during a mission~\cite{changing_number} and it is impractical to store one model per problem size on edge devices.

The mTSP class of problems are NP-hard~\cite{nphard} with the total number of possible mission plans scaling as $\frac{\left(n+m-1\right)!}{2\left(m-1\right)!}$, where $n$ is the number of tasks and $m$ is the number of agents. 
Reinforcement Learning (RL) techniques, especially those using Attention mechanisms (e.g. GAT; Graph Attention Networks), have proven to be effective at overcoming these scaling issues and efficiently solving TSP \citet{bello2017neuralcombinatorialoptimizationreinforcement,deudon, hgpn}, mTSP~\cite{rl4co}, min-max Vehicle Routing Problems (VRP)~\cite{dpn, parco}, or other similar multi-agent problems \citet{kool2019attentionlearnsolverouting, mvmoe}.

These methods are able to scale to larger problems than exact solvers such as CPLEX~\cite{cplex2009v12}, are faster at execution time and generalize better than heuristic methods~\cite{surveyrl, rl4co, surveyml} such as the Lin-Kernighan heuristic~\cite{lkh3}, show more consistent and generalizable performance than metaheuristic methods~\cite{turner_fast_2018, Li_2020}, and can be deployed on edge devices as lightweight models~\cite{8947146}.

But despite its real-world importance, no work to date has addressed the replanning problem as outlined above. Several studies have made one of the required adaptations, e.g. multiple depots, where agents start and finish at different locations~\citet{shi_multi-agent_2023,ucl_multi_robot, dpn} or cooperative tasking \citet{nazari2018reinforcementlearningsolvingvehicle}, but none have considered more than one together, and no previous studies have considered nodes with variable time costs (Table \ref{tab:gat_lit_table}), which is a fundamental requirement.
Similarly, most prior research has focused on high-performance computing systems with GPUs, with limited investigation of deployment on ground robots~\cite{ucl_multi_robot} or CPUs~\cite{nazari2018reinforcementlearningsolvingvehicle}. And whilst \cite{mvmoe, nazari2018reinforcementlearningsolvingvehicle, dpn, parco}, developed a single neural solver for different problem variants, numbers of agents or depots, they did not consider varying numbers of tasks. Our work fills this critical gap, developing a cooperative replanning solver that is suitable for edge deployment and real-world on-board application.

\begin{itemize}
    \item We are the first to formulate the Cooperative Mission Replanning Problem (CMRP); a novel and important real-world variant of the mTSP adapted to include variable agent start locations, nodes with variable time cost, and cooperative tasking.
    \item We propose a new lightweight mission replanner based on the GAT and Attention Model that can effectively and efficiently solve the CMRP on edge hardware.
    \item Using an example of UAVs completing tasks in a 2D domain, our approach produces solutions that are consistently within 10\% of those generated by the state-of-the-art LKH3 heuristic solver, but runs 85-370 times faster on a Raspberry Pi. Our single generalized model has high peformance across a range of problem sizes.
    \item Our framework also valuably supports future extension to additional important capabilities (e.g. multi-objective, probabilistic, anticipatory replanning).
\end{itemize}

Our new formulation and approach enables fast and robust in-mission, on-device replanning for the first time, for cooperative teams of robots. This has the potential to improve the resilience of diverse real-world multi-agent robotic systems.
\begin{figure*}[th!]
    \begin{center}
    \centerline{\includegraphics[width=0.8\textwidth]{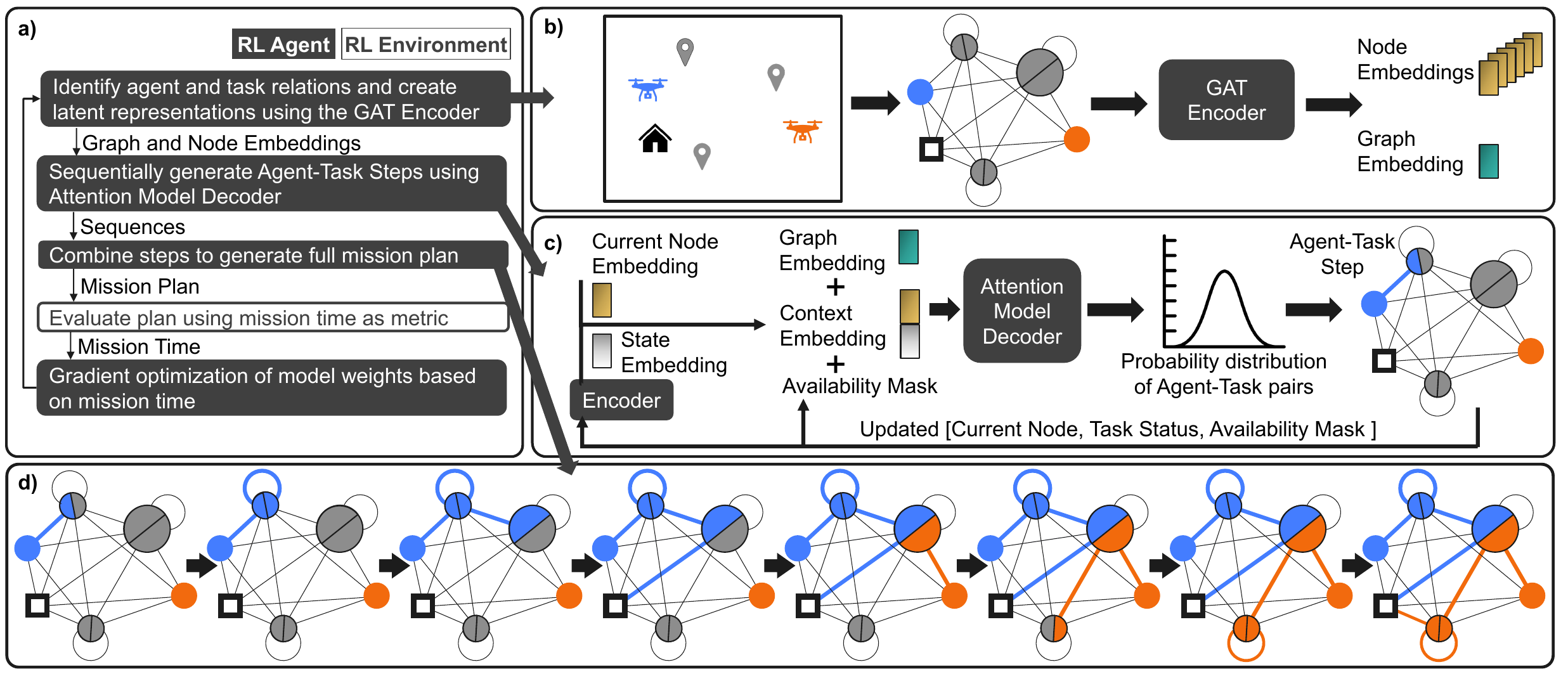}}
    \caption{(a) The RL training process. (b) The transformation of the mission planning data into higher-level node embeddings and a graph embedding using a GAT encoder. (c) The Attention Model decoder generates a probability distribution for available agent-task combinations based on the graph embedding, the current node embedding, the current state of the mission (in embedding space) and the availability mask. The agent-task step with the highest probability is then selected as the next step. (d) The sequential generation of all steps of the mission plan.}
    \label{fig:method}
    \end{center}
\vspace{-0.5cm}
\end{figure*}

\section{Methodology}
We formally define the CMRP in Section~\ref{sec:math-problem} and detail how it differs from the traditional mTSP. We then introduce our GAT-based solution to the CMRP in Section~\ref{sec:math-planner}.

\subsection{The Cooperative Mission Replanning Problem (CMRP)}
\label{sec:math-problem}
As detailed in the introduction, CMRP varies from the traditional mTSP in three key ways: 1) flexible start locations for robotic agents; 2) non-zero and variable time costs associated with each task; and 3) cooperative tasking. Additionally, there is a practical consideration of model generalization, which will be discussed later in Section~\ref{sec:generalization}.

Traditional mTSP with $m$ salespeople or agents and $n$ cities or tasks can be defined as a graph $G=(V, E)$ where the vertices $V = \{v_{depot}, v_{1}, ... v_{n} \}$ include all $n$ task locations as well as $v_{depot}$, which serves as the single start and end location for all agents. The edge set $E$ is defined as $E = \{(v_{i}, v_{j}): v_{i}, v_{j} \in V, i \neq j\}$ and represents all possible valid paths between tasks. For all edges, $T=(d_{ij})$ represents the travel time associated with traversing from task $v_i$ to $v_j$, assuming a constant travel speed of $1$~m/s.

First, we incorporate flexible start locations into this formulation by adding $m$ start locations to the vertex set, similar to the multiple depot problems \cite{shi_multi-agent_2023,ucl_multi_robot}. Our new vertex set is $V' = \{v_{depot}, v_{start1}, ... v_{startm}, v_{1}, ... v_{n} \}$, where $v_{startk}$ is the start location of the $k$-th agent, and $v_{i}$ is the location of the $i$-th task. Moreover, unlike formulations in the literature where agents complete their tours at their respective starting locations, our \( v_{depot} \) serves as a common destination for all agents, simulating a shared landing space for drones. Two constraints are set to ensure that agents always begin at their designated start locations: (1) $d_{depot\ i} = 0$; (2) $d_{i\ startj} = \infty$; where $v_{i} \in \{v_{1}, ... v_{n} \}$ and $v_{startj} \in \{v_{start1}, ... v_{startm} \}$. Hence, the $T$ cost matrix is no longer symmetric and therefore changes the problem to an asymmetric mTSP.

Second, we incorporate variable time costs for each task so that each vertex $v_{i}$ has an associated non-negative time cost $t_{i}$, with the exception of the depot and the start locations, where $t_{i}=0 \text{ for } v_{i} \in \{v_{depot}, v_{start1}, ... v_{startm} \}$. This is similar to the nodes' demand in Capacitated VRP (CVRP) \cite{kool2019attentionlearnsolverouting} with the distinction that $t_{i}$ is incorporated into the edge costs rather than being a separate vector. Our new cost matrix is therefore calculated as $T' = d_{ij} + t_{j}$, and again this modification introduces more asymmetry to this matrix, as time costs are no longer determined just by the distance between two vertices, but also by the direction of travel.

Third, each task is discretized into $\delta$ sub-tasks similar to the work in Split Delivery VRP \cite{nazari2018reinforcementlearningsolvingvehicle} to enable cooperative tasking, such that the total number of sub-tasks $n' = n \times \delta$. We refer to $\delta$ as the task discretization level. Sub-tasks replace tasks in the vertex set. They have the same spatial position as the original task but contain a corresponding fraction of the original task completion time. The time cost for a sub-task associated with task $i$ is $t'_{i} = t_{i}/\delta$. 

These combined changes from $V$, $T$, $n$, $t$ to $V'$, $T'$, $n'$, $t'$ respectively define our problem as a variant of mTSP with $m$ agents and $n' + m + 1$ vertices (including the home depot; see Fig. \ref{fig:intro}). Hence, the number of unique solutions (i.e. possible mission plans) is:

\begin{equation}
     \frac{\left(n'+m-1\right)!}{\left(m-1\right)!}
     \label{eq:scaling}
\end{equation}

\subsection{The Graph Attention Replanner (GATR)}
\label{sec:math-planner}
Our proposed GAT-based mission replanner (GATR) builds upon the encoder-decoder architecture from \citet{kool2019attentionlearnsolverouting}. We adapt the mTSP environment from the RL4CO Library \cite{rl4co} to incorporate different agents' start locations and time costs associated with nodes (the first and second changes to standard mTSP described in Table~\ref{tab:gat_lit_table}). Discretization of tasks to allow splitting of tasks between cooperative agents (the third change) does not require any additional adaptation, and can simply be accommodated during problem setup by replacing a single task of a certain time cost with several shorter collocated tasks. Finally, we enhance the data ingestion pipeline to enable training of a single generalized model for multiple problem sizes.

The overall training process is shown in Fig.~\ref{fig:method}a. Using the environment described above, we trained our model using the REINFORCE algorithm with a greedy rollout baseline \cite{kool2019attentionlearnsolverouting} (see Appendix). The reward is defined as the negative of the maximum (largest) mission time among the agents as in \cite{dpn}.

\subsubsection{Encoder}
The primary role of the GAT encoder is to transform an input graph into latent representations, as illustrated in the first step of the flowchart in Fig.~\ref{fig:method}a. In addition to the standard mTSP graph features (depot locations and task locations), we also include the time cost of vertices and agents' starting locations as part of the input to the encoder. This is similar to the approach used for specifying the `demand' for deliveries in the CVRP \citet{kool2019attentionlearnsolverouting}, and for incorporating multiple depots in \citet{shi_multi-agent_2023}. The GAT encoder then transforms these features into a set of node embeddings and a graph embedding that are passed to the decoder using the procedure described in the Appendix.

\subsubsection{Decoder}
Our decoder is based on the Attention Model, which was selected over the traditional Pointer Network due to its empirically superior performance \cite{hgpn}. The decoder forms the mission plan sequentially, as illustrated in Fig.~\ref{fig:method}c-d. At each time step, the decoder selects the next sub-task to visit for an agent based on the graph embedding, the current node embedding, the current state of the mission (in embedding space) and the availability mask. If the depot is selected, the algorithm proceeds to the subsequent agent. Note that an agent has the option to stop completing tasks and return to the depot at any point, provided there is still at least one other agent available to complete remaining tasks, similar to the assumption made in \cite{ucl_multi_robot}. The state of the environment is summarized by the following features: the number of remaining agents, the accumulated mission time for the current agent, the maximum mission time across all agents, and the distance from the depot. These features are mapped into an embedding space using affine linear transformations and concatenated with the embedding for the current agent vertex (i.e. the agent's current location) to form the context embeddings. Unlike in \citet{kool2019attentionlearnsolverouting}, we do not include the embedding of the first vertex traversed by the agent, as the agent does not return to this location in the replanning problem. Cross-attention with an availability mask is applied to the context embeddings and vertex embeddings to compute the attention scores. In the replanning problem, the mask prevents selection of: previously visited sub-tasks; start locations when the current node is not the depot; and the depot if the current agent is the last one available, and unvisited sub-tasks remain.
The next sub-task is then selected based on these scores, as detailed in the Appendix. This process is repeated until all sub-tasks are assigned to the agents. The final output is a set of routes, one for each agent, covering all sub-tasks, as illustrated in Fig.~\ref{fig:method}d. We note that in future work, the GATR architecture could be easily extended, e.g. to other encoders tailored for min-max objectives (e.g. \cite{dpn}, which has separate embeddings for task partitioning and navigation), or decoders that output actions for agents in parallel to encourage collaborative behavior \cite{parco}.

\section{Results}
\label{sec:results}
We first demonstrate GATR's effectiveness for the CMRP on small problem sizes where it is possible to evaluate our replan against all possible solutions (Section \ref{sec:small}). Then we show how a single generalized GATR model is able to solve problems with varying numbers of tasks, agents, and task discretization levels, and how our model scales to larger problem sizes (Section \ref{sec:generalization}).

\subsection{Analysis for small problem sizes}
\label{sec:small}
%\subsubsection{Experiment setup}

\begin{figure}[!ht]
    \centering
    \includegraphics[width=0.5\textwidth]{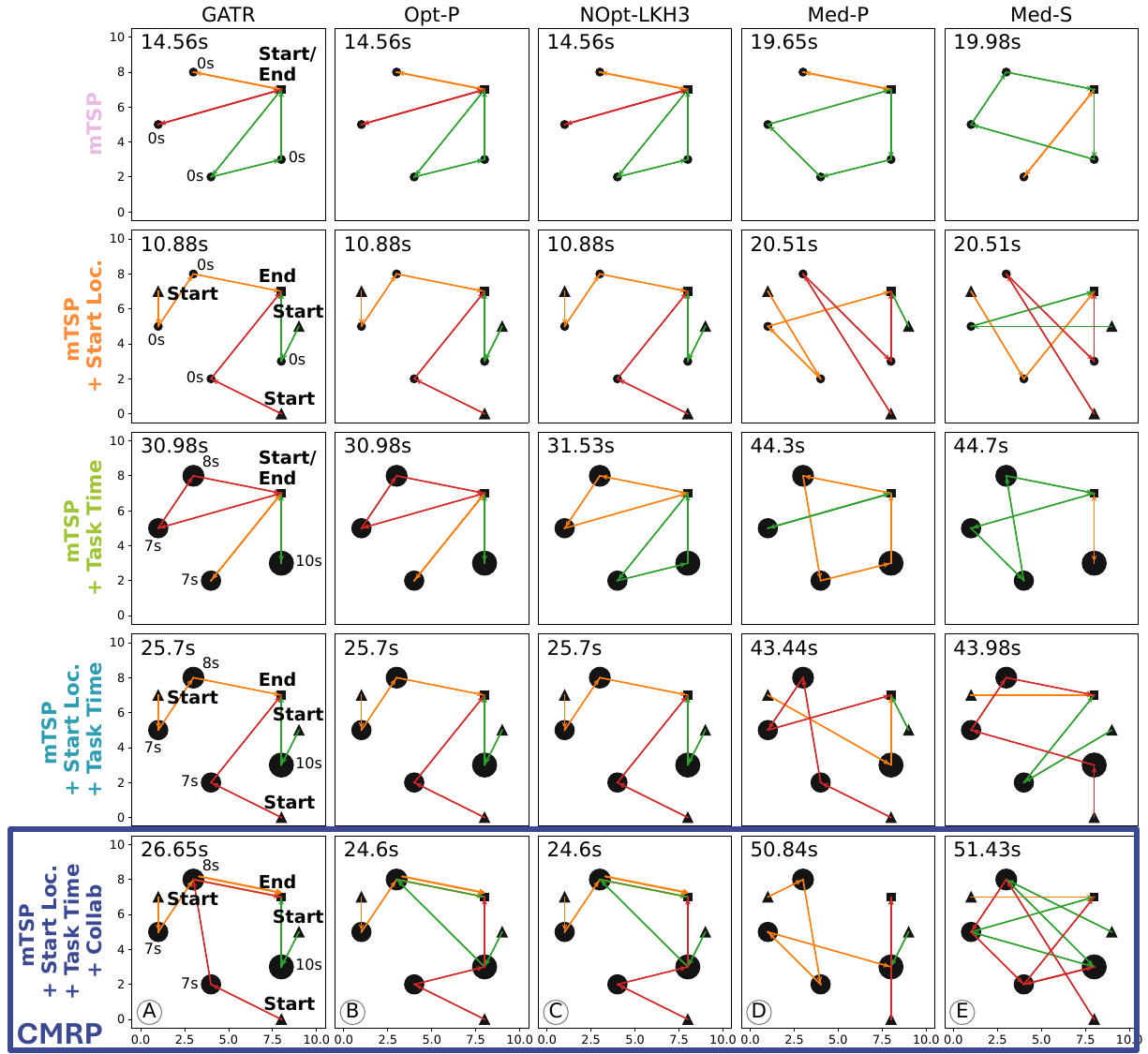}
    \caption{Comparison of how GATR and other benchmarks (columns) solve an example three-agent-four-task problem for our five increasingly complex problem types (rows). The tours of the three agents are shown in red, amber, and green, the tasks are shown in black circles (with size corresponding to the time cost of each task for the last three rows), and the black square shows the home depot. In second, fourth and fifth rows, where agent start locations differ from the depot, they are marked with a black triangle.}
    \label{fig:illustration}
\vspace{-0.4cm}
\end{figure}

\begin{figure*}[ht!]
    \centering
    \includegraphics[height=2.28in]{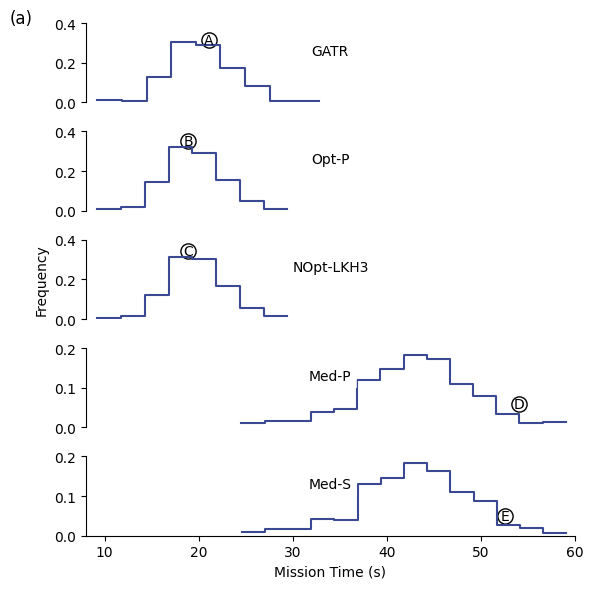}
    \includegraphics[height=2.28in]{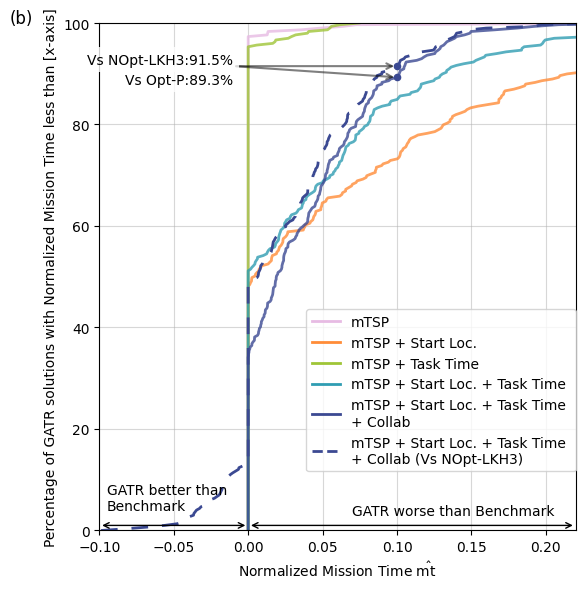}
    \includegraphics[height=2.28in]{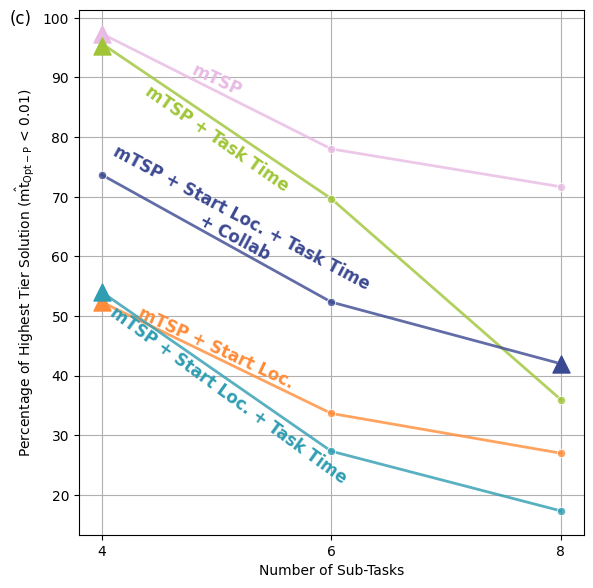}
    \caption{
 (a) Histograms showing the distribution of mission times for our method GATR (top row) and four benchmarks for the test set of 300 scenarios for the three-agent-four-task CMRP. The circled letters represent the five different solutions for the single example scenario shown in the bottom row of Fig.~\ref{fig:illustration}.
 (b) A cumulative frequency histogram showing the percentage of GATR solutions (y-axis) with normalized mission times less than a given value (x-axis). The solid lines represent comparison against the Opt-P benchmark ($\hat{mt}_{Opt-P}$) for the five progressively complex problem types illustrated in Fig. \ref{fig:illustration}, building up from mTSP (pink) to CMRP, our full problem (dark blue). The dashed dark blue line represents comparison against NOpt-LKH3 benchmark ($\hat{mt}_{NOpt-LKH3}$) for CMRP.
 (c) A graph illustrating the proportion of solutions with $\hat{mt}$ below the 0.01 (1\%) threshold for different problem types and numbers of sub-tasks. The triangles represent the four-task problems and match the colors for the same set of experiments shown in (b).}
\label{fig:scaling}
\end{figure*}

Our experimental setup simulates a small indoor flight arena for aerial drones, with a 10~m$\times$10~m 2D domain. Task, depot and agent start locations are randomly distributed, and task time costs are randomly set between 1-10~s. We assume agents move at a fixed speed of 1~m/s. In order to systematically assess GATR's capabilities, we incrementally incorporate the changes to mTSP from Section~\ref{sec:math-problem}, defining five problem types: mTSP; mTSP with multiple agent start locations; mTSP with varying task time costs; mTSP with both multiple start locations and varying task time costs; and finally our CMRP problem which also features agent cooperation (tasks discretized into multiple sub-tasks). Initially, we focus on a three agent, four task problem, and train five models (one for each problem type), using three different seeds for each. We use task discretization level $\delta=2$ for collaborative tasking, which means that our CMRP has 8 sub-tasks (4 tasks $\times 2$) whilst all other problem types have 4 sub-tasks (only 1 sub-task per task). We generate 100 mTSP evaluation instances also across three seeds (300 instances). We use mission time (time to complete all tasks and return to homebase) as our primary evaluation metric.

%\subsubsection{Benchmarks} 
We evaluate mission times against four benchmarks: the optimal and median population solutions (Opt-P and Med-P respectively), calculated via brute-force enumeration of the full population of possible solutions (e.g., 1,814,400 for 3 agents and 8 sub-tasks, from Equation \ref{eq:scaling}, evaluation of which takes $\sim$1 minute for a single scenario on the GPU); the median of a sample of the population - 101 randomly-generated solutions (Med-S); and a near-optimal solution generated by the LKH3 solver \cite{lkh3} (NOpt-LKH3). Opt-P and NOpt-LKH3 serve as the upper bounds on expected performance, whilst the two median benchmarks represent a lower bound as they simulate the expected performance of a randomly generated plan.
We note that although it is feasible to obtain enumeration solutions for these small problem sizes, this is not possible for the larger problems we investigate in Section~\ref{sec:generalization}, for which NOpt-LKH3 and Med-S are essential for evaluation instead. 

%\subsubsection{Intuitive insights}
Qualitative inspection of example solutions (e.g. single example scenario in Fig.~\ref{fig:illustration}) shows that GATR exhibits sensible and desired behaviors across all five problem types: effectively distributing tasks between agents, selecting tasks close to agent locations, accounting for variable task time cost when assigning agents, and collaboration on tasks. 

This is supported by analysis of the full test set of 300 scenarios; GATR solutions in general have mission times similar to those from Opt-P and NOpt-LKH3 solutions (Fig.~\ref{fig:scaling}a) and much smaller than those from Med-P and Med-S solutions. In addition, the close alignment between Med-S and Med-P distributions in Fig.~\ref{fig:scaling}a provides confidence in approximating the median of the full population with a sample-based estimate for evaluation of larger problem sizes.

%\subsubsection{Comparison against benchmarks}
We also analyse normalized mission time ($\hat{mt}_{benchmark}$), where GATR's mission times ($mt_{GATR}$) are compared against those from benchmark solutions:
\begin{align}
\hat{mt}_{benchmark} &= \frac{mt_{GATR}-mt_{benchmark}}{mt_{Med-P}-mt_{benchmark}}
    \label{eq:delmissiont}
\end{align}
where $mt_{benchmark}$ can refer to either mission time from the optimal benchmark ($mt_{Opt-P}$) or the near-optimal benchmark ($mt_{NOpt-LKH3}$) as desired.
Here, $\hat{mt}_{benchmark}=0$ indicates our model produces a solution with the same mission time as the chosen benchmark (the optimal solution or a near-optimal solution from the LKH3 solver); and $\hat{mt}_{benchmark}=1$ indicates our model's solution has the same mission time as an average randomly generated mission plan. As expected, optimality of the solution decreases as the problem complexity increases from mTSP (pink line in Fig.~\ref{fig:scaling}b) to the full CMRP (dark blue line). However, Fig.~\ref{fig:scaling}b shows that 89.3\% of GATR solutions still achieve $\hat{mt}_{Opt-P}<0.1$ (normalized mission times within 10\% of optimal), indicating consistently high performance despite the increased problem complexity over mTSP, and 12\% solutions actually outperform our LKH3 heuristic benchmark (evidenced by the dashed line to the left of 0.00 in Fig.~\ref{fig:scaling}b). 
%modifications are incrementally introduced to the mTSP, 
%\subsubsection{Performance scaling}
We also evaluate how performance scales with both problem complexity (Fig.~\ref{fig:scaling}b-c) and size (Fig.~\ref{fig:scaling}c). Of the two major modifications to mTSP (pink lines in Fig.~\ref{fig:scaling}b-c), the introduction of multiple start locations (orange lines) causes a larger negative impact on performance than incorporating variable task times (green lines). We suggest this is because the addition of new start location nodes causes a large increase in the number of possible solutions (as per equation \ref{eq:scaling}), whereas varying task time only affects solution cost, not the number of solutions. In terms of scaling with problem size, we run additional experiments with varying number of sub-tasks, taking the test set from 300 to 4,500 runs per method. We see an expected decrease in performance as the number of sub-tasks increases (Fig.~\ref{fig:scaling}c), but we note this trend is not strong - indeed it is only clearly visible if we analyse the proportion of highest-tier solutions (\(\hat{mt}_{Opt-P} <0.01\) in Fig.~\ref{fig:scaling}c). 
%However, our method, GATR, remains competitive at this extremely high optimality threshold. In the largest problem instance considered (8 sub-tasks), $42\%$ of solutions remain within $1\%$ deviation from the optimal solution (the dark blue triangle).

\subsection{Generalization and scaling to larger problem sizes}
\label{sec:generalization}

\begin{figure}[ht!]
    \centering
    \begin{minipage}[t]{0.4\textwidth}
        \centering
        \includegraphics[width=1\textwidth]{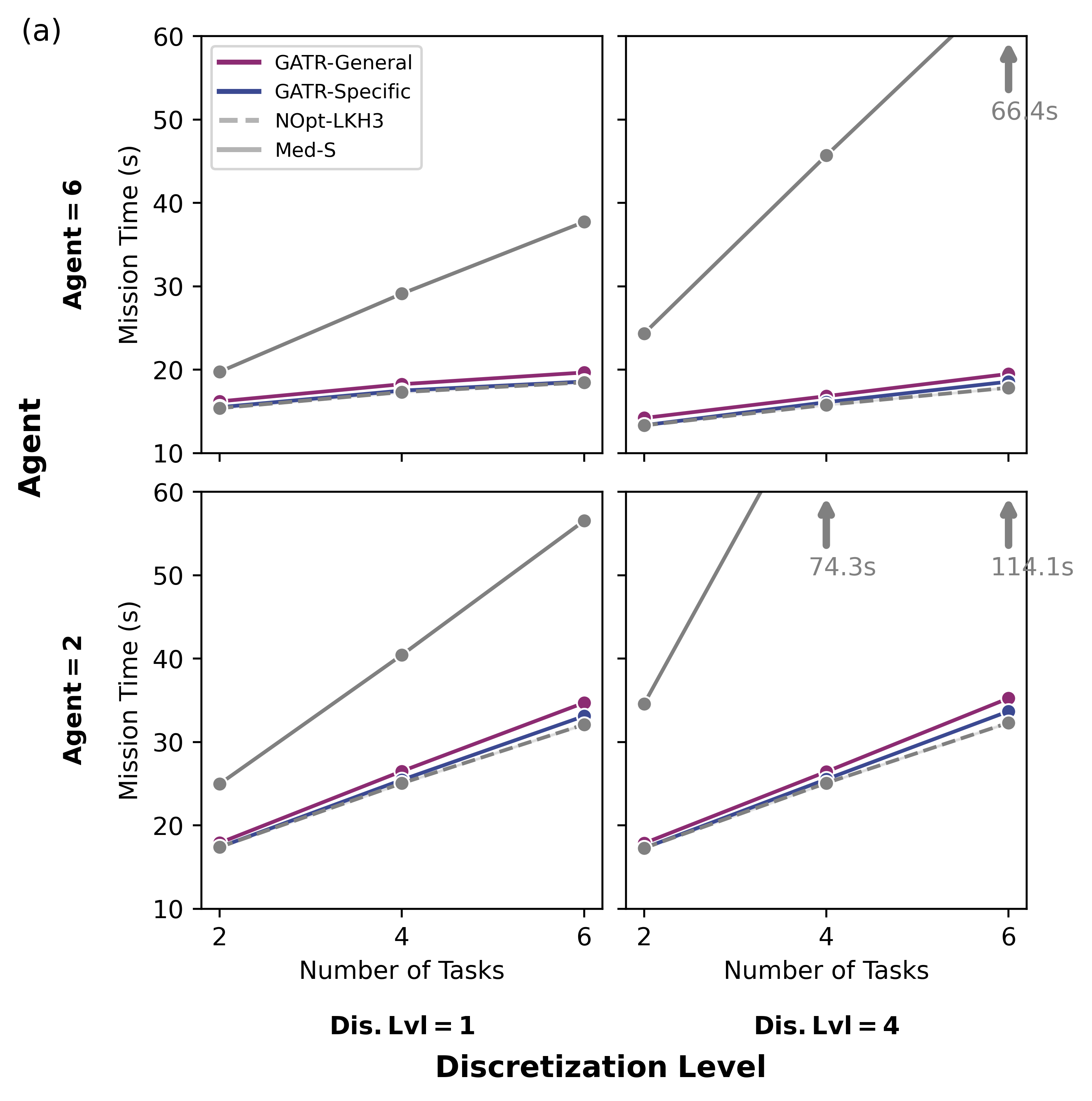}
    \end{minipage}
    \hfill
    \begin{minipage}[t]{0.4\textwidth}
        \centering
        \includegraphics[width=1\textwidth]{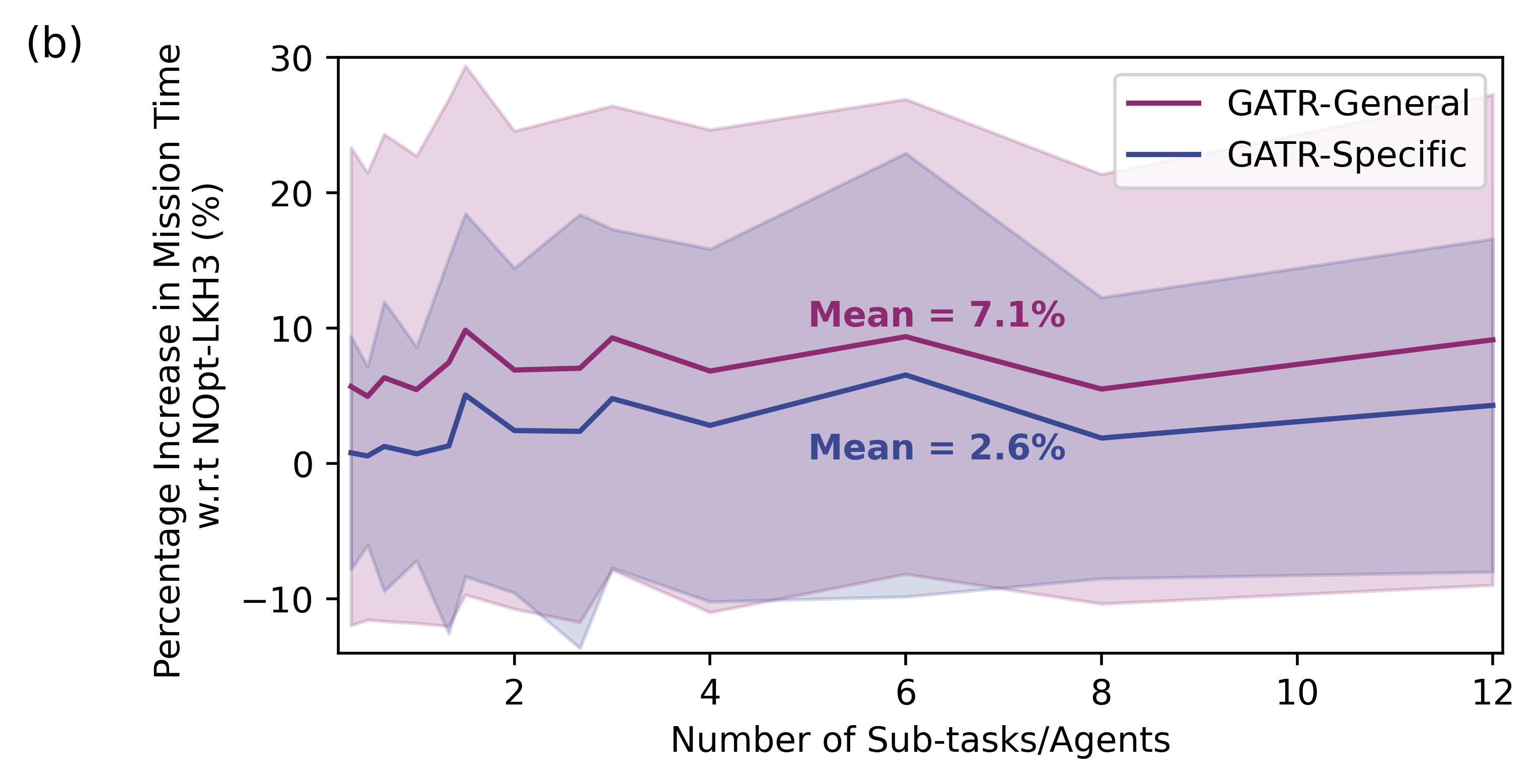}
    \end{minipage}
    \caption{
 (a) Comparison of average mission times (mean across 30,000 test scenarios) achieved by GATR-General (purple lines) and GATR-Specific (dark blue) across varying numbers of tasks, agents and task discretization levels; both closely track NOpt-LKH3 performance (dashed grey lines) and are much smaller than randomly generated missions (represented by Med-S, solid grey lines).
 (b) Comparison of percentage increase in mission time of GATR-General (purple) and GATR-Specific (dark blue) relative to LKH3 as a function of increasing problem size (number of sub-tasks per agent). Colored lines and shaded regions show mean and standard deviations respectively, with average across whole dataset indicated by colored text.
 }
    \label{fig:overall_gen}
\vspace{-0.3cm}
\end{figure}

A key implementation challenge for neural solvers is their inability to generalize to different problem setups (see penultimate column, Table \ref{tab:gat_lit_table}). But generalization is critical for replanning during a mission as the number of agents, tasks and other parameters will change throughout. This issue can be addressed by training a model on a diverse range of problem setups, but most existing implementations are constrained by fixed input graph sizes, including RL4CO that we adopt here, which supports varying numbers of agents but not tasks. To overcome this limitation, we calculate the maximum number of vertices required in the graph, i.e., $V_{max} = n_{max} \times \delta_{max} + m_{max} + 1$. Based on the scenario, we create a graph with $V = n \times \delta + m + 1$ vertices and pad it with $V_{max} - V$ vertices. These extra nodes are masked during the cross-attention procedures and will not be selected as the next action. The fixed-size matrices with custom paddings allow us to parallelize the data generation process while maintaining data diversity, hence enabling generalization. 
To demonstrate model generalization, we trained a single model (GATR-General) to solve problems with 1-6 agents, 1-6 tasks, and 1-4 task discretization levels. We tested GATR-General across 27 problem sizes (all combinations of 2/4/6 agents, 2/4/6 tasks and 1/2/4 discretization levels) and compared it against 27 specifically-trained models (GATR-Specific), one for each problem size. All models were also trained across three seeds. For fair comparison, we maintained the same network architecture to ensure similar level of RAM utilization. We set up test sets of 10,000 scenarios for each problem size with three seeds (30,000 total) as in \citet{kool2019attentionlearnsolverouting}, and evaluated GATR-General and GATR-Specific against NOpt-LKH3 and Med-S only; brute-force enumeration is not possible for large problems. Even assuming our best-case runtime for evaluating a single solution ($3\times10^{-5}\text{ s}$) from our simpler experiments (see previous Section), where it took $\sim$1 minute to obtain an optimal mission plan via enumeration, the largest problem size considered here (6 agents, 24 sub-tasks) would instead require $3\times10^{-5}\text{ s} \times \frac{(24+6-1)!}{(6-1)!} \text{ possible solutions} = 7\times10^{16} \text{ years}$. We also compared inference runtimes between GATR-General~$\text{ and}$ NOpt-LKH3 on both a high-performance GPU (single-node 40GB A100 GPU) and on edge hardware (Raspberry Pi 5 8GB), both using the same configurations (software dependencies detailed in the Appendix). We trained all models on the GPU, then transferred the resulting checkpoint files to the Pi for inference.

\begin{table}[ht!]
\caption{Runtime of GATR-General and NOpt-LKH3}
\label{tab:runtime}
\resizebox{0.5\textwidth}{!}{%
\begin{tabular}{llllllllll}
\hline
\multicolumn{1}{c}{} & \multicolumn{1}{c}{} & \multicolumn{4}{c}{4 Agents} & \multicolumn{4}{c}{6 Agents} \\
\multicolumn{1}{c}{} & \multicolumn{1}{c}{} & \multicolumn{2}{c}{6 Sub-tasks} & \multicolumn{2}{c}{24 Sub-tasks} & \multicolumn{2}{c}{6 Sub-tasks} & \multicolumn{2}{c}{24 Sub-tasks} \\
Machine & Method & Abs. & Rel. & Abs. & Rel. & Abs. & Rel. & Abs. & Rel. \\ \hline
Pi & GATR-General & 0.0078 & (1.18\%) & 0.0070 & (0.33\%) & 0.0072 & (0.67\%) & 0.0074 & (0.27\%) \\
& NOpt-LKH3 & 0.7157 &  & 2.2074 &  & 1.1301 &  & 2.7467 &  \\ \hline
GPU & GATR-General & 0.0624 & (11.11\%) & 0.0633 & (3.70\%) & 0.0726 & (7.97\%) & 0.0709 & (3.38\%) \\
& NOpt-LKH3 & 0.6138 &  & 1.7741 &  & 0.9642 &  & 2.159 & \\ \hline
\\
\multicolumn{10}{l}{\small Absolute (Abs.) times given in seconds, with relative (Rel.) times as \% w.r.t. NOpt-LKH3.} \\
\multicolumn{10}{l}{\small Standard Errors typically $<3 \times 10^{-4}$ s for GATR-General and $3-7 \times 10^{-3}$ s for NOpt-LKH3 .} \\
\end{tabular}
}
\end{table}

% using normalized runtime $\hat{rt} = \frac{rt_{GATR-General}}{rt_{NOpt-LKH3}} \times 100\%$.
%\subsubsection{Generalization Performance}
Overall, both GATR-General and GATR-Specific closely follow LKH3's performance across a range of problem sizes (Fig.~\ref{fig:overall_gen}a). GATR-General and GATR-Specific achieve mission times that are only 7.1\% and 2.6\% larger than LKH3 respectively, and this gap does not show strong signs of widening with increasing workload per agent (Fig.~\ref{fig:overall_gen}b, purple and dark blue lines). Furthermore, GATR-General's runtime is just 0.27--1.18\% of LKH3's runtime on edge hardware (e.g. 0.007~s vs. 2.7~s at lower end, see first two rows in Table \ref{tab:runtime}); this fast speed is essential for real-time on-board replanning, especially when many individual replans may be required across multiple agents, e.g. for consensus-based planning (see Discussion). Table \ref{tab:runtime} (compare first two rows) shows GATR-General maintains consistent runtimes across varying numbers of sub-tasks from 6 to 24 (similar to \cite{shi_multi-agent_2023}), whereas LKH3's runtime scales poorly, increasing by roughly a factor of four. Moreover, execution on the Raspberry Pi is faster than on the GPU because there is limited room for parallelization when replans are only required in isolation.

%Moreover, the performance gap between GATR-General and GATR-Specific remains stable across problem sizes, indicating robust generalisation. Additionally, normalized runtime $\hat{rt}$ decreases as workload per agent increases, likely because GAT models shift task allocation from an all-at-once to a sequential, point-by-point approach. Consequently, their runtime remains nearly constant regardless of problem size \citet{shi_multi-agent_2023}. In contrast, LKH3's runtime by default scales with the number of vertices, explaining the linear trend in Figure~\ref{fig:scaling}. Interestingly, due to the streaming mode, there is limited room for parallelization, making execution on the Raspberry Pi faster than on the GPU.
\section{Discussion}
\label{sec:discussion}
 Our GATR framework enables effective and efficient replanning for cooperative multi-agent robotic systems, but also has the major advantage of enabling future extension to incorporate additional key capabilities; multi-objective, probabilistic consensus, and anticipatory replanning.

\subsubsection{Managing competing aims - multi-objective planning}
In this study, our GATR method is designed to minimize overall mission time only, which implies that there are no practical constraints on maximum overall mission time or on fuel usage for each robotic agent, and that all tasks are expected to be completed. 
However, these assumptions are not valid for many real-world scenarios, where the ability to consider and explore trade-offs between mission completion time, task completion rate (percentage of tasks completed), and agent attrition rate (percentage of agents lost, e.g. due to insufficient fuel for return to homebase) is important.
Existing heuristic or meta-heuristic methods are not designed to handle such multi-objective planning, but our encoder-decoder GATR framework can easily support adaptation to incorporate these considerations in future work, primarily through modifying our model's reward function to include terms for task completion rate and agent attrition rate, but also through modification of the model's state space (e.g. adding elements to represent agent fuel levels, sensor functionality, actuator health), asymmetric edge weights and node costs (e.g. to represent wind or currents, which influence agent traversal and task completion times). These adaptations would also allow replanning for response to mission environment changes (e.g. reduction in mission time window due to incoming storm or imposition of a no-fly zone, change in travel times due to increased wind \cite{palma}) or more granular changes in system health (e.g. reduction of agent speed \cite{Ure}), in addition to responding to the changes already demonstrated in this study; i.e. in tasking (changes in location, number, time-cost of tasks) and high-level system health (loss or addition of agents).

\subsubsection{Managing current uncertainty - probabilistic consensus planning}
To develop our GATR method in this study, we have assumed for simplicity that all robotic agents have the same perfect (i.e. zero uncertainty) knowledge of the environment state. Under these idealized conditions, all agents will generate the same replan for any given scenario; we therefore do not explicitly consider which agent generates any replan. But in real-world dynamic environments with limited communication, agents' observations of the environment will be subject to uncertainty, and these observations and uncertainties will differ between agents \cite{heter_agent}. Both of these challenges can be addressed through future incorporation of probabilistic Bayesian methods into GATR, which is straightforward for RL-based models \cite{bayesian_rl} such as ours, but not for meta-heuristic or heuristic planning methods. The lightweight and fast nature of GATR (runtime order of 0.07~s) makes it feasible to use such a probabilistic approach for consensus replanning, by generating replans from multiple robotic agents with differing beliefs. This consensus replanning could be achieved via a proposer-responder framework, in which any agent within a subset can temporarily assume the role of leader \cite{dynamic_leader}, propose a plan, and allow other agents to vote based on their own beliefs about the global state, merging multiple beliefs to reduce input uncertainties and achieving replanning in a semi-decentralized fashion. The combination of diverse beliefs and independent planning fosters ensemble learning, where collective intelligence can be used to enhance the overall planning quality \cite{bayes_bots}. Our GATR framework can therefore serve as a fundamental building block for development of a probabilistic-guided, reliable and resilient semi-decentralized replanner.

\subsubsection{Managing future uncertainty - anticipatory planning}
Our GATR method provides the ability to replan onboard, which is essential for adapting to unforeseen mission changes \cite{live-fly}. But if we have knowledge of the type of changes and their likelihood, then anticipatory planning can enable our replans to mitigate the impact of further changes.
The effectiveness of GATR could therefore be further enhanced by integrating anticipatory planning techniques. This could be achieved through further leveraging Bayesian methods to construct beliefs not just about neighbouring agents' positions and task progress \cite{bayesian_rl}, but also about the likelihood of future changes that might necessitate a replan, e.g. to system health (failure of agents, loss of fuel), tasking (addition of new tasks, changes in tasks), or to the environment (reduction of overall mission time). For instance, there is a significant likelihood of path deviations when wind speeds are high, or sensors may degrade more rapidly in hazardous environments, preventing agents from completing assigned tasks \cite{Ure}. Replanning based on probabilistic beliefs of further changes, rather than assuming deterministic execution of the replan, would enable the system to regulate minor deviations before they escalate into agent failures or mission-wide breakdowns \cite{replan_bayes}. This proactive error regulation would reduce the frequency of replanning and the magnitude of required deviations from the initial plan or from any replans, leading to more efficient overall mission execution. 
\subsubsection{Conclusion}
In this study we have defined the Cooperative Mission Replanning Problem (CMRP), which represents a key real-world challenge for developing resilient, autonomous multi-robot systems, and have developed GATR, a new model to solve this problem effectively and efficiently. This work enables on-board replanning for cooperative multi-robot systems for the first time, improving their adaptability and resilience in dynamic, uncertain environments.

\addtolength{\textheight}{-0cm}   
\appendix
\noindent \textbf{Graph Attention Model Overview:} Following \citet{attentionisallyouneed}, we use query/key dimensionality $d_{k} = \frac{\text{Hidden Dimensions}}{\text{Attention Heads}} = \frac{128}{8} = 16$.

\textit{Encoder:} The features of the input graph's vertices (coordinates and time cost) are represented as $\mathbf{x_{i}} = [x_{i}, y_{i}, t_{i}]$. These are transformed into embeddings: $h_{i}^{(0)} = Linear(\mathbf{x_{i}})$. The embeddings $h_{i}^{(0)}$ are processed through $N$ attention layers to generate vertex embeddings~$h_{i}^{(N)}$, and the graph embedding is computed as $\bar{h}^{(N)} = \frac{1}{n} \sum_{i=1}^{n} h_{i}^{(N)}$.

\textit{Decoder:} The state embedding is defined as $\mathbf{s} = [m_{remaining}, t_{cur}, t_{max}, dist]$ and transformed into $h_{state} = Linear(\mathbf{s})$. The context embedding combines the graph embedding with the previous vertex embedding: $h_{(c)}^{(N)} = [\bar{h}^{(N)}, Linear([h_{state}, h_{\text{pre. vertex}}^{(N)}])]$. The probability of selecting a vertex is computed as: $p_{i} = \frac{e^{u_{i}}}{\sum_{i} e^{u_{j}}}$, where $u_{j} = -\infty$ if $j \in$ Mask; otherwise $u_{j} = 10 \cdot \tanh \left(\frac{q' \cdot k'_{j}}{\sqrt{d_{k}}} \right)$, where queries $q'$ and keys $k'$ are products of learnable weights and embeddings.

\noindent \textbf{Training:} We train with a batch size of 5000 and a learning rate of $1 \times 10^{-4}$ for up to 50 epochs with early stopping. Using 1,000,000 training and 100,000 validation samples, model training on an NVIDIA A100 GPU takes $\sim$1 hour. Major software dependencies include: Python 3.10.16, RL4CO 0.5.0.dev, TorchRL 0.6.0, PyTorch-Lightning 2.3.8.

\bibliography{bibtex/bib/references}{}
\bibliographystyle{IEEEtran}

\end{document}